\documentclass[letterpaper, 10 pt, conference]{ieeeconf}  

\IEEEoverridecommandlockouts                              

\usepackage{graphicx}
\usepackage{wrapfig}
\usepackage{eucal}
\usepackage{color}
\usepackage{amsmath}
\usepackage{lipsum} 
\usepackage[T1]{fontenc}
\usepackage{newtxtext}
\usepackage[cmintegrals]{newtxmath}
\usepackage{bm} 
\usepackage[linesnumbered,ruled,vlined]{algorithm2e}
\usepackage{subfigure}
\usepackage[utf8]{inputenc}
\usepackage{graphicx}
\usepackage{graphics}
\usepackage{siunitx}
\usepackage{multirow}
\usepackage{float}
\usepackage{caption}
\usepackage{gensymb}
\usepackage{hyperref}
\usepackage{epstopdf}
\usepackage{enumerate}
\usepackage{multicol,multirow}
\usepackage{booktabs}
\usepackage{cite}
\usepackage{graphicx}
\usepackage{textcomp}
\usepackage{xcolor}
\usepackage{wrapfig}
\usepackage{amsmath}
\usepackage{cleveref}

\usepackage{paralist}

\usepackage{graphicx}
\usepackage{wrapfig}
\usepackage{eucal}
\usepackage{color}
\usepackage{amsmath}
\usepackage{lipsum} 
\usepackage[T1]{fontenc}
\usepackage{newtxtext}
\usepackage[cmintegrals]{newtxmath}
\usepackage{bm} 
\usepackage[linesnumbered,ruled,vlined]{algorithm2e}
\usepackage{subfigure}
\usepackage[utf8]{inputenc}
\usepackage{graphicx}
\usepackage{graphics}
\usepackage{siunitx}
\usepackage{multirow}
\usepackage{float}
\usepackage{caption}
\usepackage{gensymb}
\usepackage{hyperref}
\usepackage{epstopdf}

\usepackage{multicol,multirow}
\usepackage{booktabs}
\usepackage{textcomp}
\usepackage{xcolor}
\usepackage{wrapfig}
\usepackage{amsmath}
\usepackage{cleveref}

\graphicspath{{}}

\newcommand{\vect}[1]{\boldsymbol{\mathit{\bm{#1}}}}

\definecolor{red}{rgb}{1.00,0.00,0.00}
\definecolor{lightred}{rgb}{1.00,0.3,0.3}
\definecolor{blue}{rgb}{0.00,0.00,1.00}
\definecolor{green}{rgb}{0.1,0.50,0.1}
\definecolor{yellow}{rgb}{0.5,0.5,0.0}
\definecolor{white}{rgb}{1,1,1}
\definecolor{gray}{rgb}{0.6,0.6,0.6}

\newcommand{\cblue}[1] {\textcolor{blue}{#1}}

\usepackage{array, makecell, rotating}
\setcellgapes{2pt}
\newcommand\nocell[1]{\multicolumn{#1}{c|}{}}

\title{\LARGE \bf
  A Synergistic Framework for Learning Shape Estimation and Shape-Aware Whole-Body Control Policy for Continuum Robots}

\author{Mohammadreza Kasaei$^{1}$, Farshid~Alambeigi$^{2}$, and Mohsen~Khadem$^{1}$
\thanks{$^{1}$ Mohammadreza Kasaei and Mohsen Khadem are with the School of Informatics, University of Edinburgh, UK. Email: \{m.kasaei, mohsen.khadem\}@ed.ac.uk}%
\thanks{$^{2}$F.~Alambeigi is with the Walker Department of Mechanical Engineering, the University of Texas at Austin, Austin, TX, USA. Email: farshid.alambeigi@austin.utexas.edu}
\thanks{
This work is supported by the Medical Research Council [MR/T023252/1]}
}

\newenvironment{customindent}[1]%
  {\begin{list}{}%
     {\setlength{\leftmargin}{#1}}%
     \item[]%
  }
  {\end{list}}

\begin{document}

\maketitle
\thispagestyle{empty}
\pagestyle{empty}

\begin{abstract}
In this paper, we present a novel synergistic framework for learning shape estimation and a shape-aware whole-body control policy for tendon driven continuum robots. Our approach leverages the interaction between two Augmented Neural Ordinary Differential Equations~(ANODEs) --- the \textit{Shape-NODE} and \textit{Control-NODE} --- to achieve continuous shape estimation and shape-aware control. The \textit{Shape-NODE} integrates prior knowledge from Cosserat rod theory, allowing it to adapt and account for model mismatches, while the \textit{Control-NODE} uses this shape information to optimize a whole-body control policy, trained in a Model Predictive Control~(MPC) fashion. This unified framework effectively overcomes limitations of existing data-driven methods, such as poor shape awareness and challenges in capturing complex nonlinear dynamics. Extensive evaluations in both simulation and real-world environments demonstrate the framework’s robust performance in shape estimation, trajectory tracking, and obstacle avoidance. The proposed method consistently outperforms state-of-the-art end-to-end, Neural-ODE, and Recurrent Neural Network~(RNN) models, particularly in terms of tracking accuracy and generalization capabilities. The code and pretrained models are available at \cblue{\small{\url{https://github.com/SIRGLab/WholeBodyControl_CTR}}}.
\end{abstract}

\section{Introduction}
\label{sec:intro}
Soft continuum robots are constructed from flexible materials such as rubber, silicone, or elastomers, allowing them to conform to surfaces and objects while maintaining a level of durability that rigid robots do not possess~\cite{laschi2016soft,rus2015design}. This adaptability makes them well-suited for diverse applications, including medical procedures~\cite{sharma2023novel, 10335932, mklung}, search and rescue operations~\cite{yamauchi2022development}, and exploration missions~\cite{wooten2018exploration}. Despite these advantages, designing and controlling soft continuum robots remain challenging~\cite{george2018control, review2}. These challenges stem from the difficulty of predicting their shapes due to nonlinear behaviors and the need to model complex structural deformations with high degrees of freedom, making precise whole-body control a particularly complex task~\cite{mengaldo:hal-03921606}.


\subsection{Shape Estimation}
The shape of continuum robots is typically described by kinematic or dynamic models that relate the configuration of the robot's backbone to its joint inputs. Various mathematical approaches have been developed for modeling continuum robots, including Piecewise Constant Curvature~(PCC)~\cite{webster2010design,katzschmann2019dynamic}, polynomial curvature fitting~\cite{della2019control}, lumped parameter models~\cite{della2018dynamic,della2020model}, and reduced-order finite element models~\cite{duriez2013control,katzschmann2019dynamically}. A detailed review of these methodologies is provided in~\cite{review3}. One of the most widely used frameworks for modeling continuum robots is Cosserat rod theory~\cite{till2019real, rucker2, mohsen1}. However, shape estimation based on model-driven approaches often suffers from significant inaccuracies due to the dependence on external force information, which is often unavailable~\cite{mackute2022shape}. Additionally, the inherent challenges posed by highly non-linear kinematics, torsion, and friction further compromise the accuracy of these models~\cite{xu2017data}. Furthermore, these methods can be computationally intensive and may not adequately represent the complex nonlinear dynamics inherent to soft robots. 
\begin{figure}[!t]
    \centering
    \includegraphics[width=\linewidth]{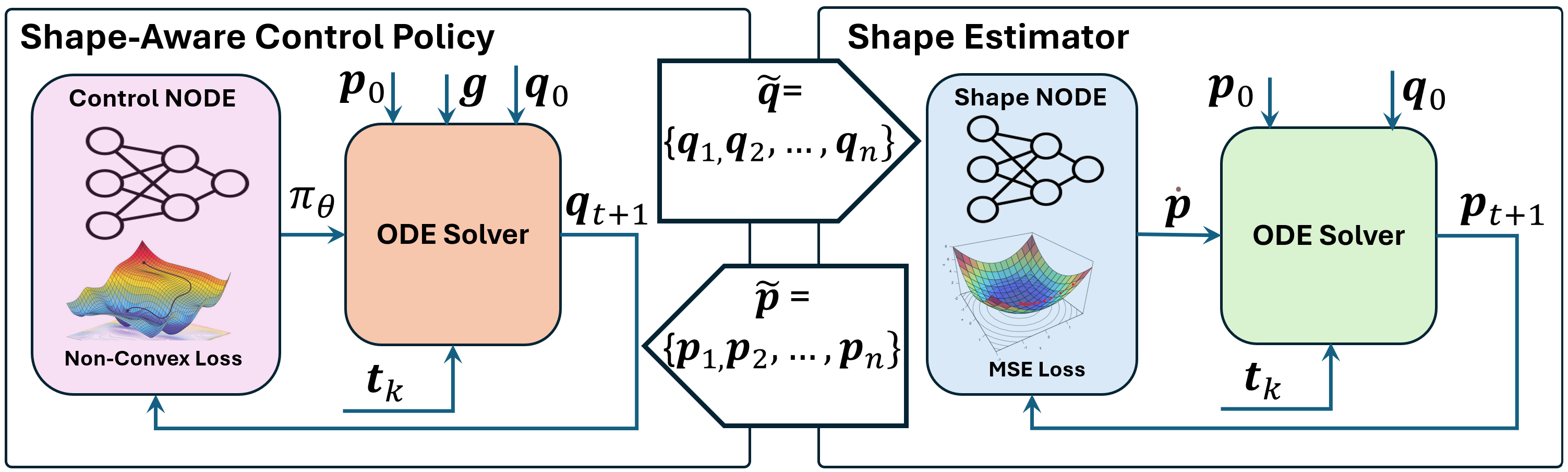}
    \caption{Overall architecture of the proposed framework, consisting of two Augmented Neural ODEs: Shape-NODE and Control-NODE. The former is for learning a continuous shape estimation and the latter for learning a shape-aware whole-body control policy.}
    \label{fig:overview}
    \vspace{-3mm}
\end{figure}

Data-driven approaches for continuum robot modeling leverage collected data to develop models and controllers, either independently or alongside mathematical models~\cite{gao2024sim}. These learning-based methods are generally more resilient to assumptions from physics-based models and less affected by fabrication errors, particularly when trained directly on the physical robot. Although data-driven methods offer the potential to address the limitations of model-based approaches, they come with challenges. They require large amounts of training data, which can be difficult to collect and may risk damaging the robot during the process. Additionally, these methods often struggle to generalize to unforeseen scenarios and lack interpretability, making it harder to understand the robot's control policy and physical behavior.

\subsection{Control}
Controlling continuum robots is particularly challenging due to their geometric and behavioral non-linearities. Traditional physics-based models, such as 
PCC approximation, struggle with capturing the full complexity of these systems and are often computationally expensive. Continuum mechanics based models such as Cosserat rod theory help model continuum structures, but simplifying assumptions can reduce accuracy, especially in accounting for nonlinear deformation of continuum structures. New approaches, such as data-driven models~\cite{balint1, balint2,bern2020soft,thuruthel2017learning,bruder2019nonlinear,morimoto2021model,buchler2022learning}, aim to learn kinematics or control policies using real robot data, offering improved performance. However, these methods require extensive data and suffer from limitations like lack of generalization to unseen environments. Despite their promise, further advancements are needed to develop models that account for the complex dynamics of soft continuum robots. Recent studies, such as~\cite{kasaei2023data2,kasaei2023CORL}, utilize Neural Ordinary Differential Equations~\cite{chen2018neural,dupont2019augmented} to learn continuous non-parametric kinematic models of continuum robots, followed by the implementation of traditional PD controllers or Model Predictive Control~(MPC) to control the robot's tip motion. However, these approaches overlook the robot's overall shape, making them unsuitable for scenarios where robot interacts with environment and shape awareness is critical.

\subsection{Contributions}
In this paper, we introduce a novel synergistic framework for learning both shape estimation and a shape-aware whole-body control policy for continuum robots. A key aspect of our approach is framing the problem as the interaction of two Augmented Neural Ordinary Differential Equations (ANODEs), referred to as the Shape-NODE and Control-NODE. The Shape-NODE is responsible for learning continuous shape estimation, and its formulation incorporates prior knowledge from Cosserat rod modeling, allowing it to account for and learn model-mismatch. The Control-NODE leverages the shape information to optimize a whole-body control policy, which is trained in an MPC fashion. This collaboration between the Shape-NODE and Control-NODE enables the framework to overcome limitations of previous data-driven approaches, such as poor shape-awareness and the inability to capture complex nonlinear dynamics. The framework has been fully tested in both simulated environments and on a real continuum robot, validating its ability to perform robust, accurate, shape-aware whole-body control.

The remainder of this paper is organized as follows: Section~\ref{sec:method} details the proposed methodology, including the design and training of the Shape-NODE and Control-NODE. In Section~\ref{sec:sim}, we present simulation results to evaluate the performance of the proposed framework in various scenarios.
Section~\ref{sec:exp} discusses the experimental setup and results from real-robot tests. Section~\ref{sec:comp} presents a comparison of the framework with existing approaches. Finally, conclusions are discussed in Section~\ref{sec:conclusion}.

\section{Methodology} \label{sec:method}
This section explains detail of our proposed methodology.
We begin by presenting an overview of the proposed framework, followed by detailed discussion of the shape estimator and the shape-aware whole-body control policy.

\subsection{Overview}
Figure~\ref{fig:overview} presents an overview of the proposed framework, which comprises two ANODEs: the \textit{Shape-ANODE} and the \textit{Control-ANODE}. The \textit{Shape-NODE} is designed to learn a continuous shape estimation function, while the \textit{Control-NODE} is developed to learn a shape-aware whole-body control policy. Both ANODEs are trained holistically to optimize their joint performance.

\subsection{Shape-NODE: Continues Shape Estimator Function}
Cosserat rod theory has been widely used to represent the
geometry of continuum robots, which assigns a
material frame along the center-line curve such that the z-axis of the frame is tangent to the curve. The homogeneous
rigid-body transformation \( \mathbf{T}(s,t) \in \text{SE}(3) \) is used to describe the
evolution of the position and orientation of the frame:
\begin{equation}
\mathbf{T}(s,t) = 
\begin{bmatrix}
\mathbf{R}(s,t) & \vect{p}(s,t) \\
\mathbf{0}_{1 \times 3} & 1
\end{bmatrix},
\end{equation}
where \(\vect{p}(s,t): [0, \ell] \times [0, \infty) \rightarrow \mathbb{R}^3\) and \(\mathbf{R}(s,t): [0, \ell] \times [0, \infty] \rightarrow \mathit{SO}(3)\) are position and orientation, respectively, \( s \in \mathbb{R} \) is the arc-length parameter, and \(\ell\) represents the length of the robot. The shape of the main backbone can be characterized by the following equations \cite{rucker2}:
\begin{equation}
\mathbf{T}'(s,t) = \mathbf{T}(s)[\xi(s,t)]_\times, \quad
\vect{u}'(s,t) = h(s, \vect{u}, q(t)),
\label{3}%
\end{equation}
where \( \xi(s) = (\vect{u}(s), e_3) \in \mathbb{R}^6 \) is the body twist of the material frame,  \( (\cdot)' \) denotes the partial derivative with respect to arc-length $s$, $[.]{{\times}}$\ denotes the \( 4 \times 4 \) matrix representation of
twist \cite{mrbook}, \(\vect{u}(t) = [u_x(t), u_y(t), 0]^T\) represents the curvature vector of the deformed backbone, and \( e_3 \in \mathbb{R}^3 \) is the z-directional unit vector. $h(s, \vect{u}, q(t))$ is a nonlinear function of robot arclength, curvature, and actuation inputs $q(t)$. Shape of the robot can be found by solving the above ODE for any arbitrary point along robot arclength $ S \in [0, \ell] $ at time $t$ as

\begin{equation}
\vect{p}(S,t) = \int_0^{S} \vect{p}(s,t) \text{d}s,
\label{eq:math_model}%
\end{equation} 
\noindent
using this notation, the cartesian coordinates of robot tip is 
$\vect{x}(t)= \vect{p}(\ell,t) = \int_0^{\ell} \vect{p}(s,t) \text{d}s$.
In the case of a multi-segment continuum robot, each segment is characterized by its own set of centroids and transformation matrices. To represent the complete configuration of the robot, the output of the integration for each segment serves as the initial condition for the subsequent segment, ensuring a continuous and smooth deformation across the entire robot structure.

Based on the mathematical modeling described, we conceptualize the shape estimation problem as an ANODE. This framing enables the network to be initialized using the mathematical model and subsequently adapt to mismatches between the model and actual data through training. Let \( f \) represent a \textit{nonlinear stiff differential equation}, which is distinguished by its solutions exhibiting both rapidly and slowly varying components. This function \( f \) can encapsulate the entire shape of the robot in Cartesian coordinates as follows:

\begin{equation}
\begin{aligned}
&{\vect{p}}^{'}(s,t) = f(\vect{p}, \vect{u}(t)), \\
& f: \mathbb{R}^3 \times \mathbb{R}^3 \rightarrow \mathbb{R}^3
\end{aligned}
\label{ode}
\end{equation}
with initial conditions \(\vect{p}_0\) and \(\vect{u}_0\). To approximate~(\ref{eq:math_model}), we can utilize a time-dependent multilayer perceptron~(MLP) to solve the initial value problem~(IVP), \mbox{\(\frac{\partial \vect{p}(t)}{\partial t} = f_{\theta}(\vect{p}(t), u(t), t)\)}, and compute the shape using a numerical ODE solver that solves the ODE for a fixed timestamp $t$:
\begin{equation}
     \hat{\vect{p}}_{S_{i+1}} = \operatorname{ODESolver}(f_{\theta}, (\vect{p}_{S_{i}}, \vect{u}_{({S_{i}}.t)}), (S_{i}, S_{i+1})),
\end{equation}
here, we drop \( t \) for simplicity, as the equations are solved along the arc length \( s \) for a fixed \( t \).
By discretizing the model described in Equation~(\ref{ode}), we convert it into a boundary value problem~(BVP):
\begin{equation}
\vect{p}^{+} = f_{\theta}(\vect{p}(S), \vect{u}(S)),
\label{ode2}
\end{equation}
subject to the boundary conditions:
\begin{equation}
\begin{aligned} 
\vect{p}_0 = \vect{p}(s=0), \quad \vect{u}_0 = \vect{u}(s=0), \\
\vect{p}_k = \vect{p}(s=S_k), \quad 
\vect{u}_k = \vect{u}(s=S_k).
\end{aligned}
\label{BC}
\end{equation}

Given that the neural network, \(f_{\theta}\), approximates the function \(f\), we can compute the solution \( \vect{p}(s=S_k) \) in case we have knowledge of \(f_{\theta}\):

\begin{equation}
    \vect{p}(S_{k+1}) = \vect{p}(S_{k}) + \int_{S_{k}}^{S_{k+1}} f_{\theta} (\vect{p}(s), \vect{u}(s)) \, ds,
    \label{eq:h}
\end{equation}
\noindent
thus, standard numerical ODE solvers like the Euler method, Runge-Kutta method, or fixed-Adams method can be employed to approximate \(\vect{p}(S_{k})\):
\begin{equation}
    \hat{\vect{p}}(S_{k+1}) = \text{ODESolver}(f_{\theta}, (\vect{p}(S_{k}),\vect{u}(S_{k})), (S_{k}, S_{k+1})).
    \label{eq:ode_solver}
\end{equation}
However, if \(f_{\theta}\) is inaccurate or unknown, we can use the error to train the network:

\begin{equation}
    \text{Loss}_1 = \Vert \hat{\vect{p}}(S_k) - \vect{p}(S_k) \Vert .
    \label{loss}
\end{equation}

Equation~(\ref{eq:ode_solver}) allows for estimating the robot's shape across a batch of samples simultaneously \((\vect{u}^N = \{\vect{u}_0, \vect{u}_1, \ldots, \vect{u}_N\})\) but requires a consistent integration length for all samples. Since the integration length, determined by length of robot \(\ell(t)\) in Equation~(\ref{eq:math_model}), varies with input conditions, the maximum length is calculated across the batch. The system is then uniformly evaluated up to this maximum length. The ODE is solved simultaneously for all samples, and solutions beyond each sample's actual length are masked by retaining the last valid state for subsequent steps, ensuring accurate output. Details of the training parameters and network architecture are provided in Table~\ref{tab:hyper}.


\subsection{Control NODE: Shape-Aware Whole-Body Control Policy}
After training the \textit{Shape-NODE}, to train a shape-aware whole-body control policy, $\pi_\theta$, that leverages the \textit{Shape-NODE}, we employed another ANODE that takes as input the shape information~\mbox{\((\vect{P}^N = \{\vect{p}_0, \vect{p}_1, \ldots, \vect{p}_N\})\)}, the current tip position of the robot~($\vect{x}_t$) at time $t$, current actions~($\vect{q}_t$), and the desired trajectory~($\vect{g}_t$) to generate a sequence of actions \mbox{\((\vect{q}^M = \{\vect{q}_0, \vect{q}_1, \ldots, \vect{q}_M\})\)} over a predefined finite horizon, $M$, subject to the \textit{Shape-NODE}. The mathematical formulation is as follows:
\begin{equation}
\begin{aligned}
     &\vect{q}^M = \operatorname*{ODESolver}(\pi_{\theta}, \vect{p}_0, \vect{x}_0, \vect{q}_0,  \mathbf{g}, (t_0, \ldots, t_{M-1})), \\ 
     &\text{For} \;\; \vect{q}_t \in \vect{q}^M, \;\; \vect{p}_{S+1} = \operatorname*{Shape-NODE}(\vect{p}_S, \vect{u}_S), \\
     &\text{where}, \;\;  \vect{p}_S = 
     \begin{cases}
         \vect{p}_0, & \text{if} \;\; s=0 \\
         \operatorname*{Shape-NODE}(\vect{p}_{S-1}, \vect{u}_{S-1}), & \text{if} \;\; s > 0
     \end{cases}
\end{aligned}
\end{equation}

The training process for the \textit{Control-NODE} in a multi-segment continuum robot involves iteratively optimizing the policy for precise and smooth movement. At the start of each epoch, the robot's initial states are randomized by applying perturbations to a set of reset actions, which act as the initial inputs to the robot. These randomized actions are passed to the \textit{Shape-NODE} to simulate the robot's initial states, which can be downsampled to reduce complexity, serving as input for the \textit{Control-NODE}.

Next, target tip positions are generated by perturbing the final tip positions from the initial state for $M$ samples. This introduces variability in the desired end-effector positions, challenging the controller to adapt to a range of conditions. The neural network processes the downsampled shape information, current tip position, current actions, and target positions to generate the necessary actions to move the robot towards the targets. These actions are then fed into the \textit{Shape-NODE}, simulating the robot's movement and resulting in a sequence of states that describe its configuration over time.

In our training approach, the policy is optimized in an MPC fashion, predicting a sequence of~\(M\)~actions that guide the robot's tip along a desired trajectory, subject to constraints imposed by the action bounds. The actions generated by the policy are constrained within predefined limits to ensure feasible and safe control of the robot. Specifically, the action values are scaled using the \texttt{Tanh} activation function and mapped to a desired range using minimum and maximum action values. The loss function combines the errors in the tip trajectory, action regularization, shape consistency, and terminal cost into a single expression:
\begin{figure*}[!t]
    \centering
    \subfigure[]{
        \includegraphics[width=0.23\textwidth]{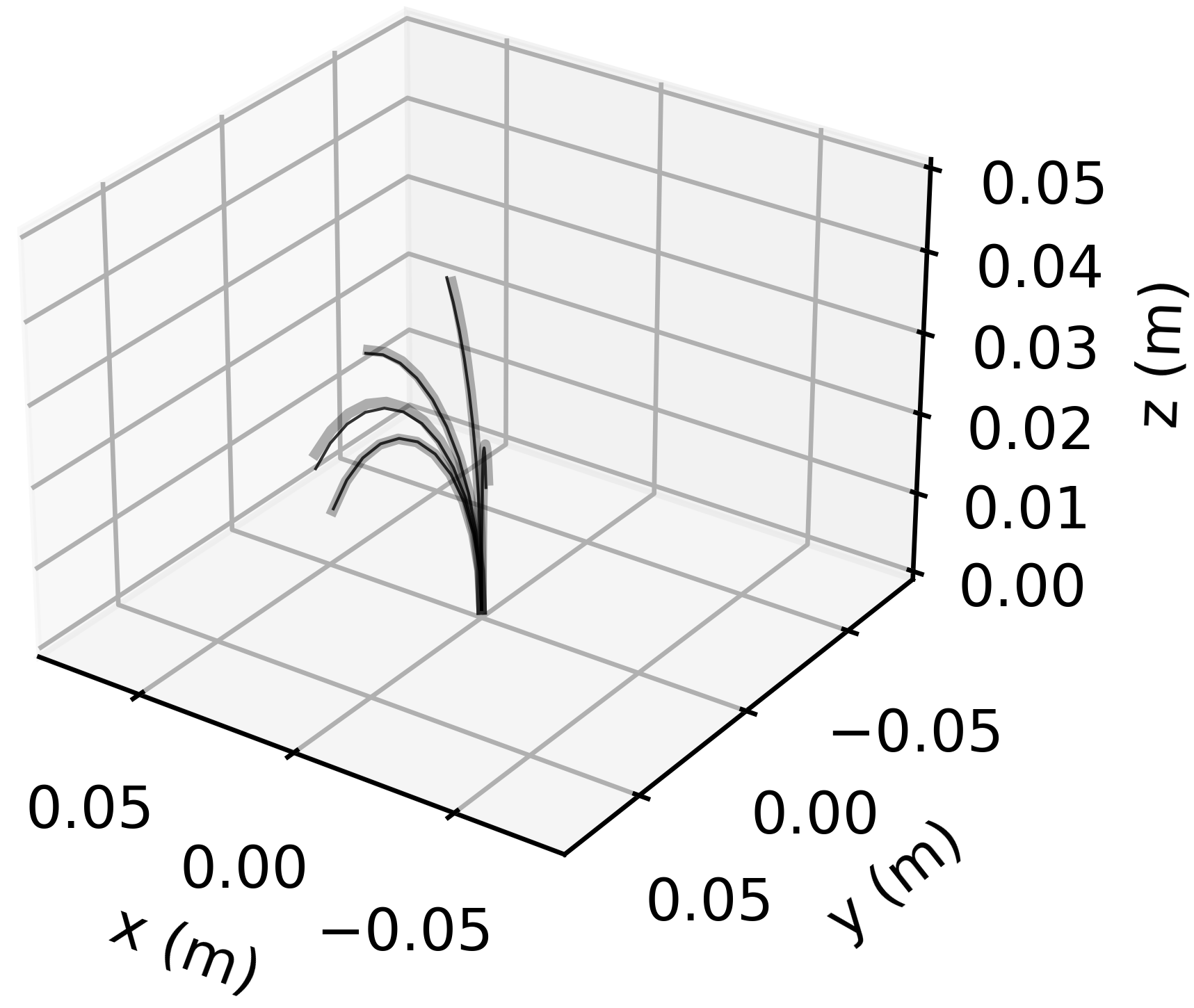}
    }
    \subfigure[]{
        \includegraphics[width=0.23\textwidth]{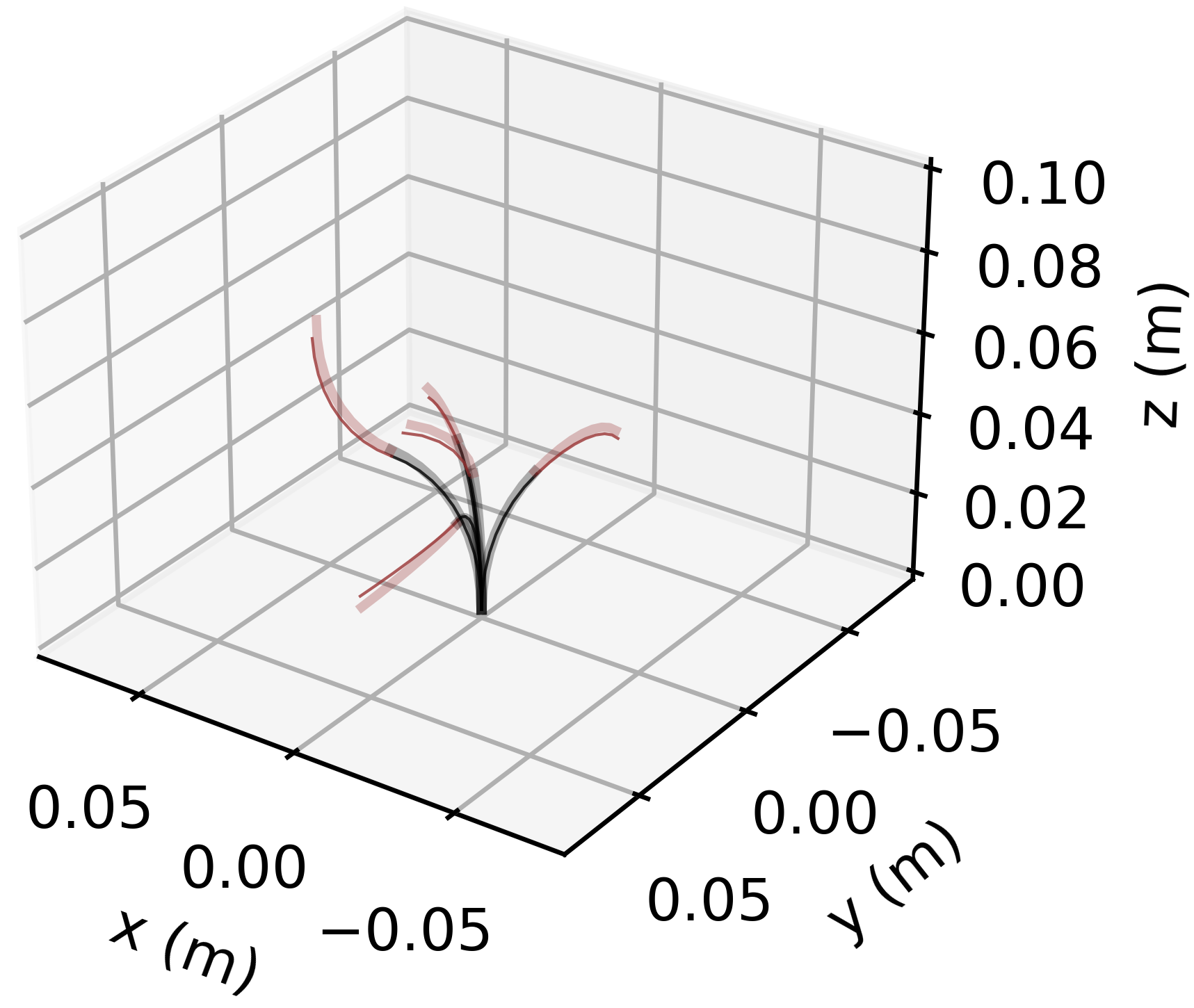}
    }
    \subfigure[]{
        \includegraphics[width=0.23\textwidth]{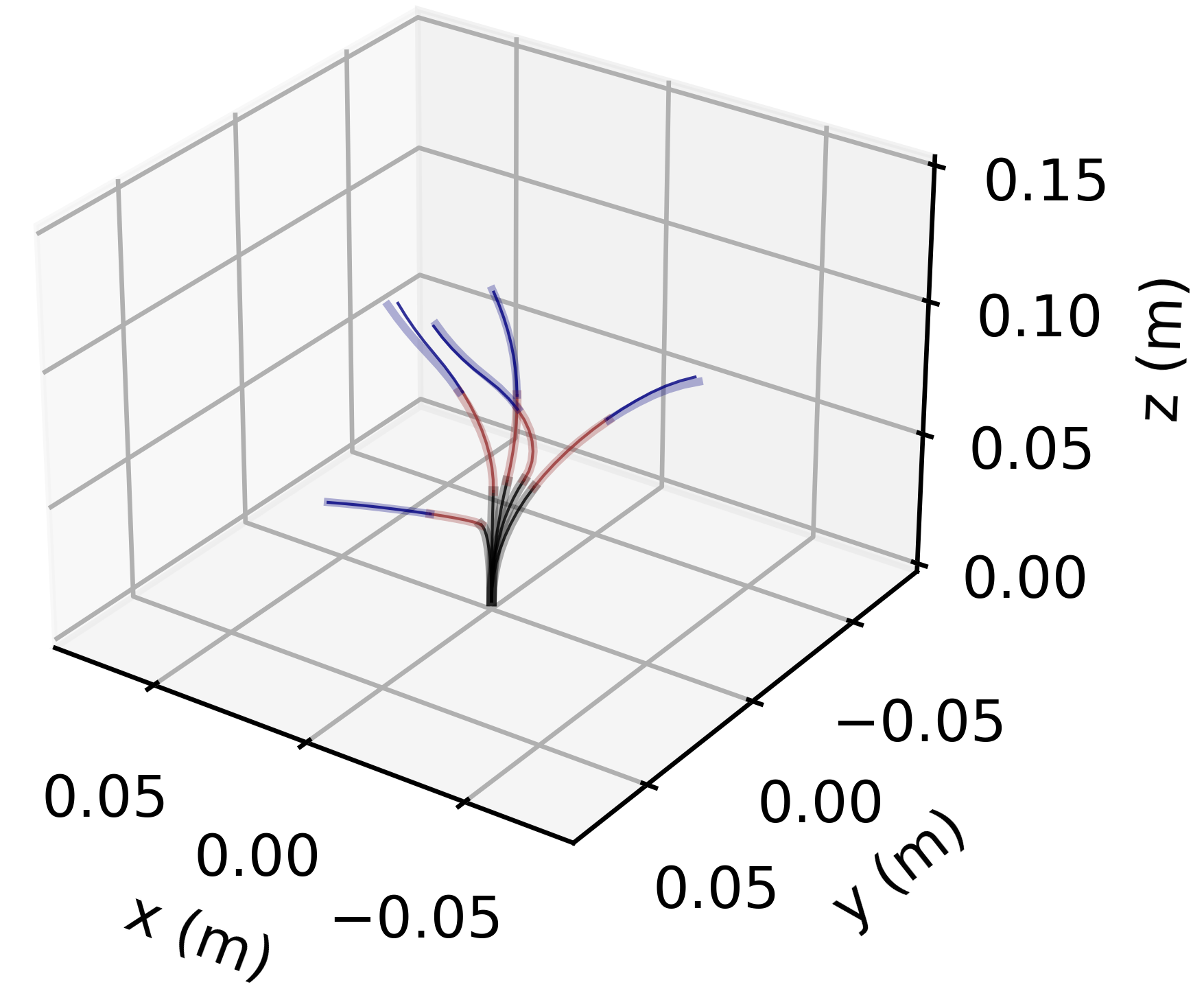}
    }
    \subfigure[]{
        \includegraphics[width=0.23\textwidth]{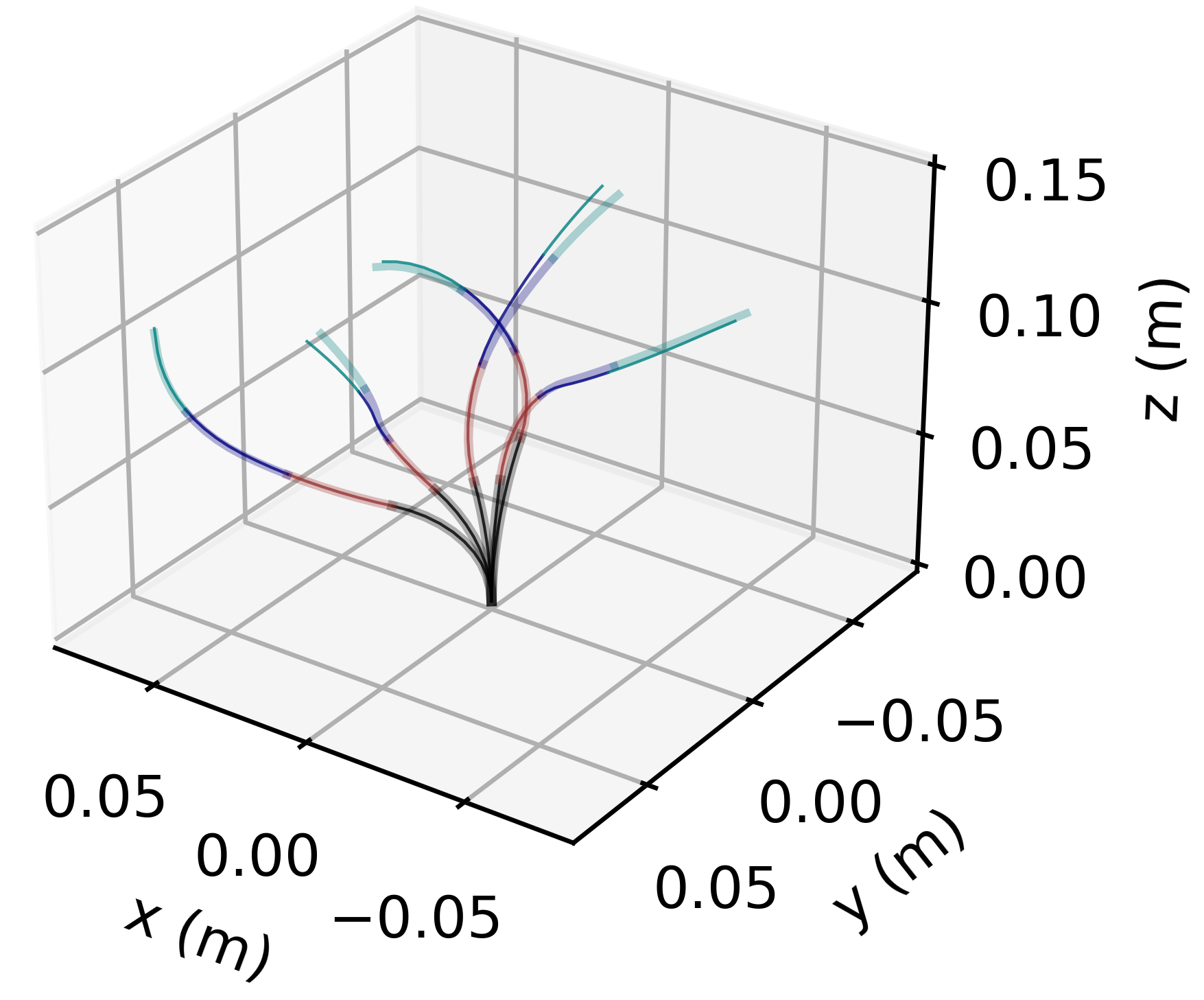}
    }
    \vspace{-3mm}
    \caption{Representative results for the shape prediction scenario (five prediction per robot): solid lines indicate the ground truth, while the transparent lines depict the predicted shapes. The cases include (a)~a single-segment robot, (b)~a two-segment robot, (c)~a three-segment robot, and (d)~a four-segment robot.}
    \label{fig:shape}
    \vspace{-8mm}

\end{figure*}

\begin{equation}
\begin{aligned}
\text{Loss}_2 = & \sum_{k=1}^{M}  \Big( \alpha \|\hat{\vect{x}}(t_k) - \mathbf{g}(t_k)\|^2  +  \beta \|\hat{\vect{q}}(t_k) - \vect{q}(t_{k-1})\|^2  + \\
& \gamma \|\hat{\vect{p}}(S_k) - \vect{p}(S_{k-1})\|^2 \Big) + \lambda \|\hat{\vect{x}}(t_M) - \mathbf{g}_(t_M)\|^2,
\end{aligned}
\label{eq:loss}
\end{equation}
\noindent
where \(\hat{\vect{x}}(t_k)\) and \(\mathbf{g}(t_k)\) are the predicted and target tip positions at time step \(t_k\), respectively; \(\hat{\vect{q}}(t_k)\) and \(\vect{q}(t_{k-1})\) are the predicted and initial actions; \(\hat{\vect{p}}(S_k)\) and \(\vect{p}(S_{k-1})\) are the shape information predicted and initial shape information. The coefficients \(\alpha\), \(\beta\), \(\gamma\), and \(\lambda\) weight the importance of minimizing trajectory errors, ensuring smooth actions, maintaining state consistency, and minimizing the terminal cost at the final time step \(t_M\), respectively. To have a more robust policy, we add a Gaussian noise to the observation during the training ($\mathcal{N}(0,0.00033^2)$). It is important to note that additional objectives, such as obstacle avoidance, can be incorporated into the loss function if needed. 

\begin{table}
\vspace{-2mm}
\caption{Model Hyperparameters and Training}
\resizebox{0.48\textwidth}{!}{
\begin{tabular}{ |c | c | c |}
    \cline{2-3}
    \nocell{1} &  \multicolumn{2}{|c|}{\textbf{Hyper-parameters}}  \\ \cline{2-3}
    \nocell{1} & \textbf{Control NODE} & \textbf{Shape NODE}  \\                 
 \hline
    architecture   & 39@256@256@6* & 7@256@256@7*   \\ \hline 
    activation   & LeakyReLU(@) ; Tanh(*) & LeakyReLU(@) ; Tanh(*)  \\    \hline
    optimizer & Adam & Adam   \\    \hline
    learning rate  & 1e-3 & 1e-3 \\\hline
     batch size & 256 & 256\\ \hline
     total iterations & 10k & 10k \\ \hline
     ODE solver & fixed-adams & fixed-adams \\ \hline
      $\alpha$, $\beta$, $\gamma$, $\lambda$  &5000; 100; 200; 1000 & - \\ \hline
\end{tabular}}
\label{tab:hyper}
\vspace{-0mm}
\end{table}

\section{Simulations}
\label{sec:sim}
In this section, we will design and conduct a series of simulation scenarios using SoftManiSim~\cite{kasaeisoftmanisim} --- a simulation framework that combines advanced continuum robot modeling with the PyBullet simulator~\cite{coumans2020} --- to evaluate the performance of the proposed framework.

\begin{customindent}{-3pt}
\begin{itemize}
    \item \textbf{Shape Estimation Evaluation:} This simulation scenario evaluates the performance of \textit{Shape-NODE} in predicting the shapes of multi-segment continuum robots, ranging from one to four segments. We generate a shape dataset for these robots using SoftManiSim, then train and assess \textit{Shape-NODE}'s ability to accurately estimate their shapes.
    \item \textbf{Trajectory Tracking:} The goal of this simulation is to evaluate the performance of \textit{Control-NODE}. Utilizing \textit{Shape-NODE}s trained in the previous scenario, we train corresponding \textit{Control-NODE}s for various trajectory tracking, including a circular shape with a radius of 0.05~meters, a square with side lengths of 0.06~meters, and S-shapes defined by the equations \(x = a \cos\left(\frac{2\pi t}{T}\right)\) and \(y = b \sin\left(\frac{4\pi t}{T}\right)\), where \(a = 0.03\), \(b = 0.05\), and \(T = 100\) seconds. The parameter \(t\) varies between 0 and 100 seconds, and ellipses.
    \item \textbf{Obstacle Avoidance:} The objective of this simulation is to validate the performance of the proposed framework in enabling a three-segment continuum robot to track different trajectories while avoiding a static obstacle.
\end{itemize}
\end{customindent}

\begin{figure*}[!t]
    \centering
    \subfigure[]{
        \includegraphics[width=0.23\textwidth]{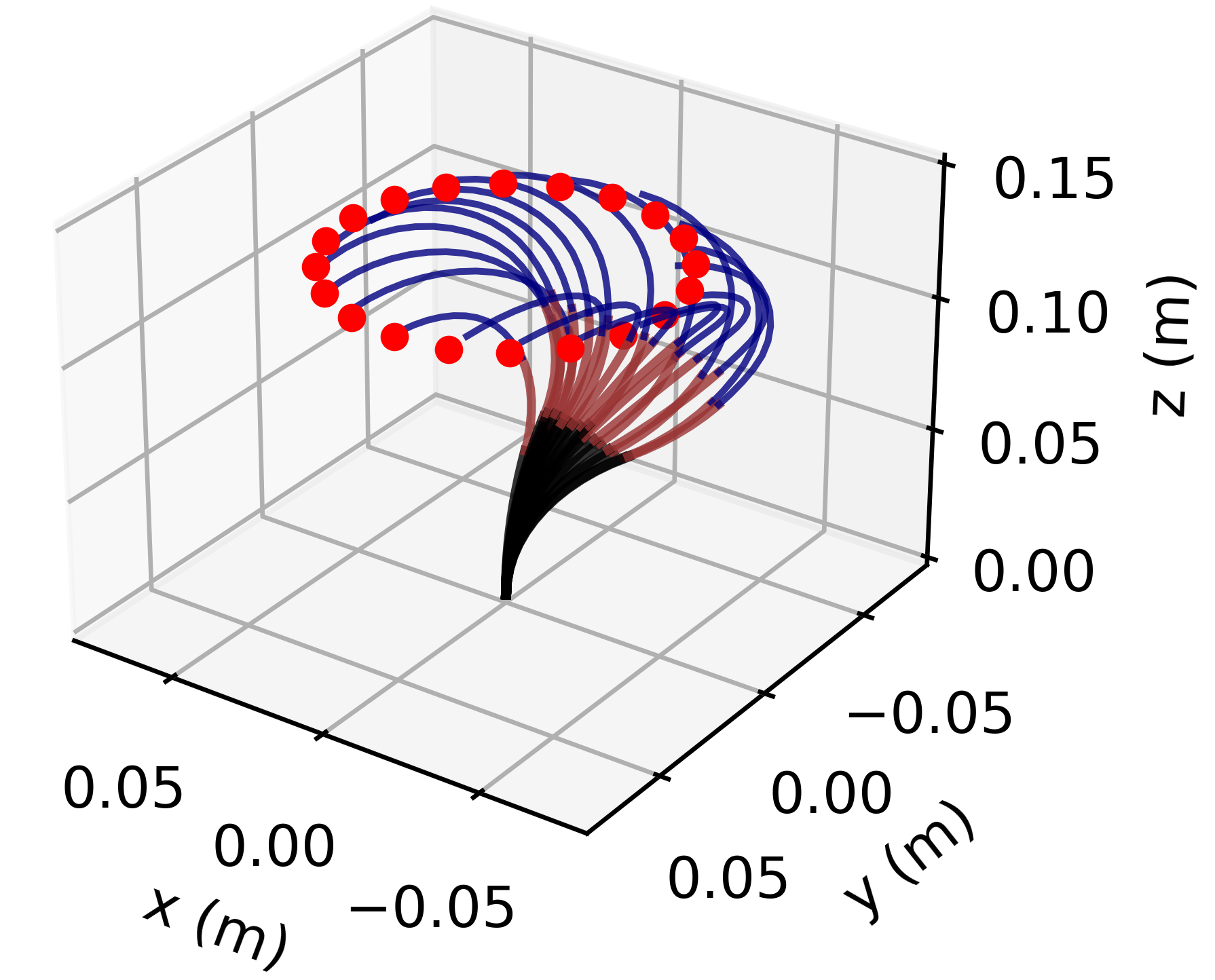}
    }
    \subfigure[]{
        \includegraphics[width=0.23\textwidth]{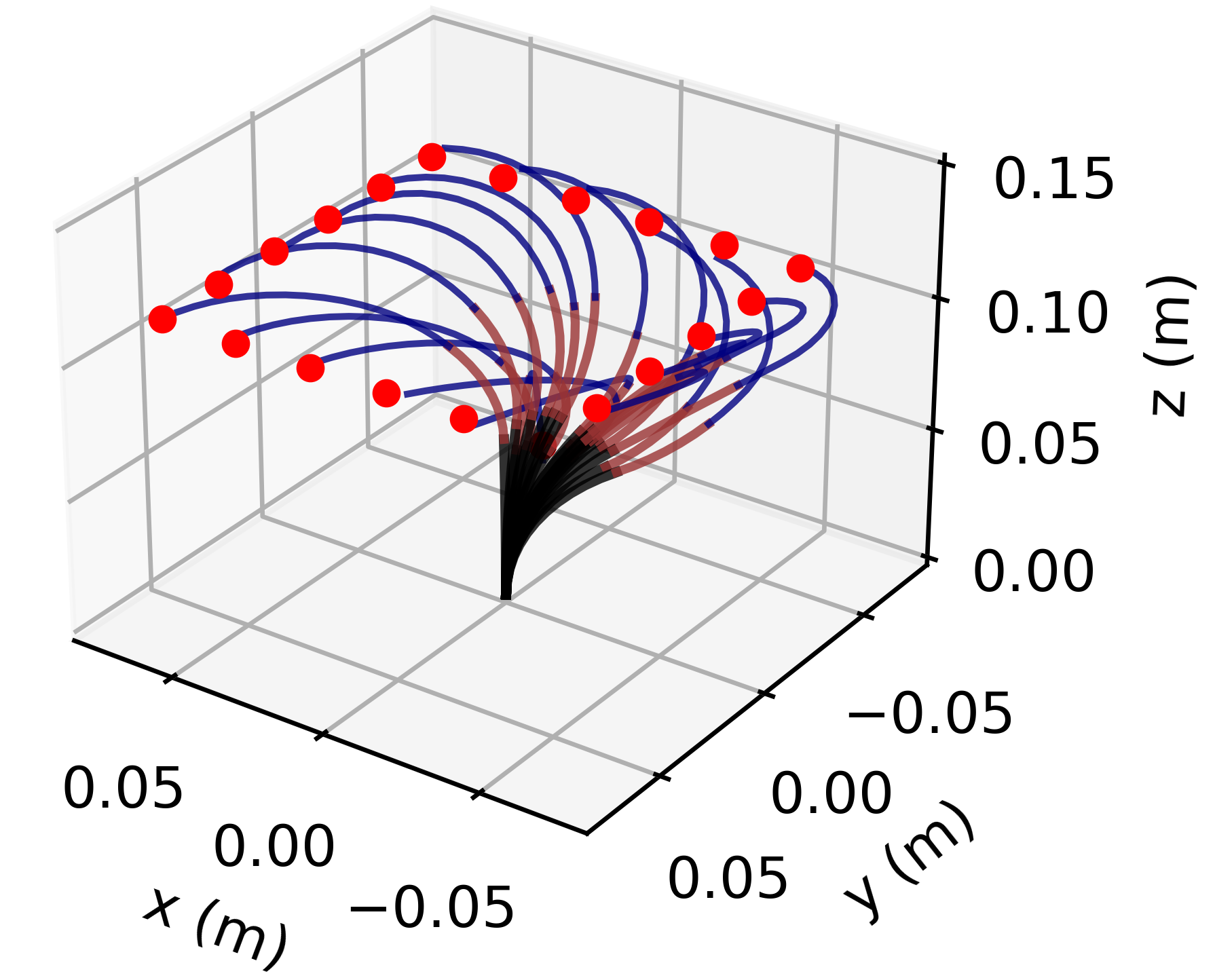}
    }
    \subfigure[]{
        \includegraphics[width=0.23\textwidth]{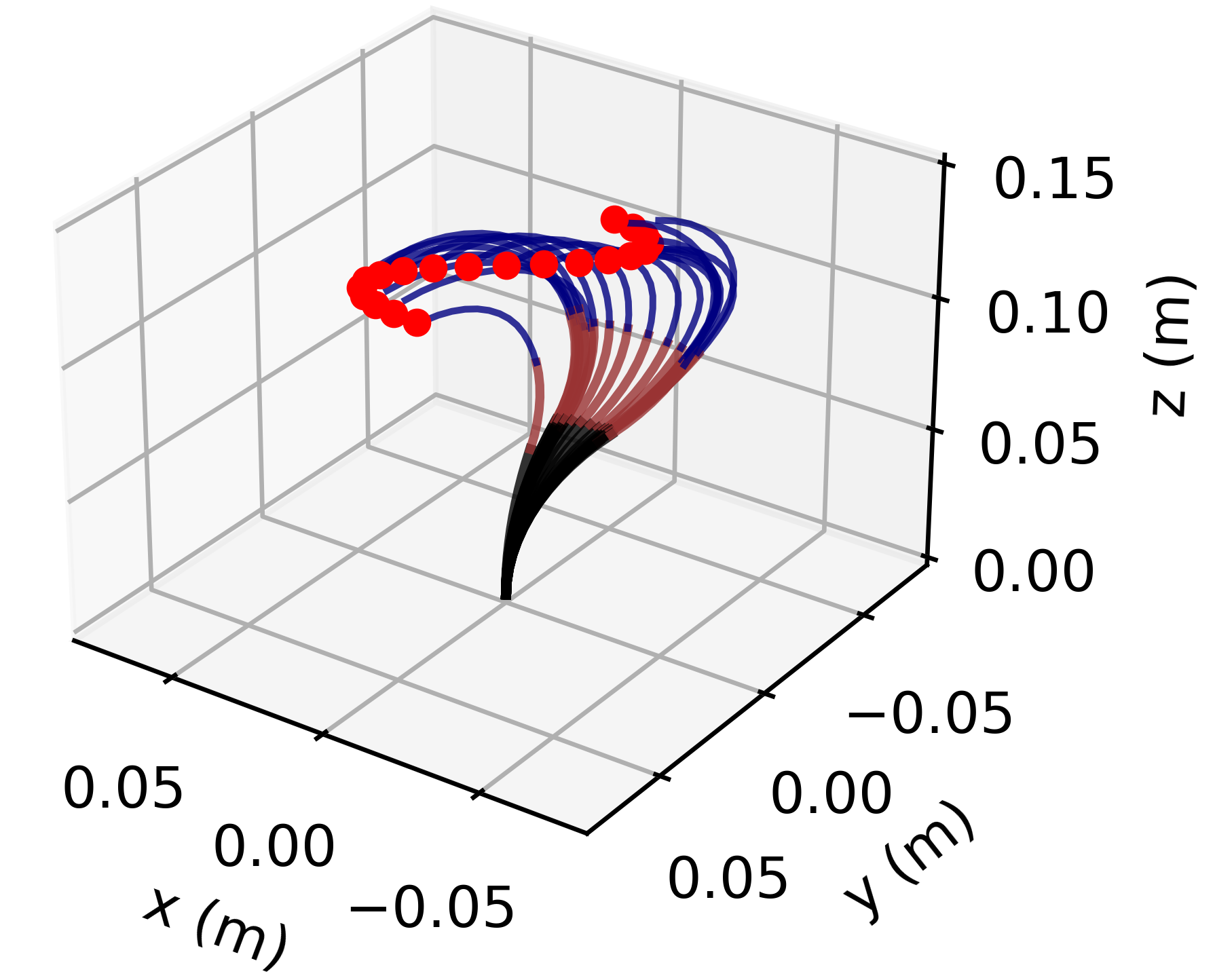}
    }
    \subfigure[]{
        \includegraphics[width=0.23\textwidth]{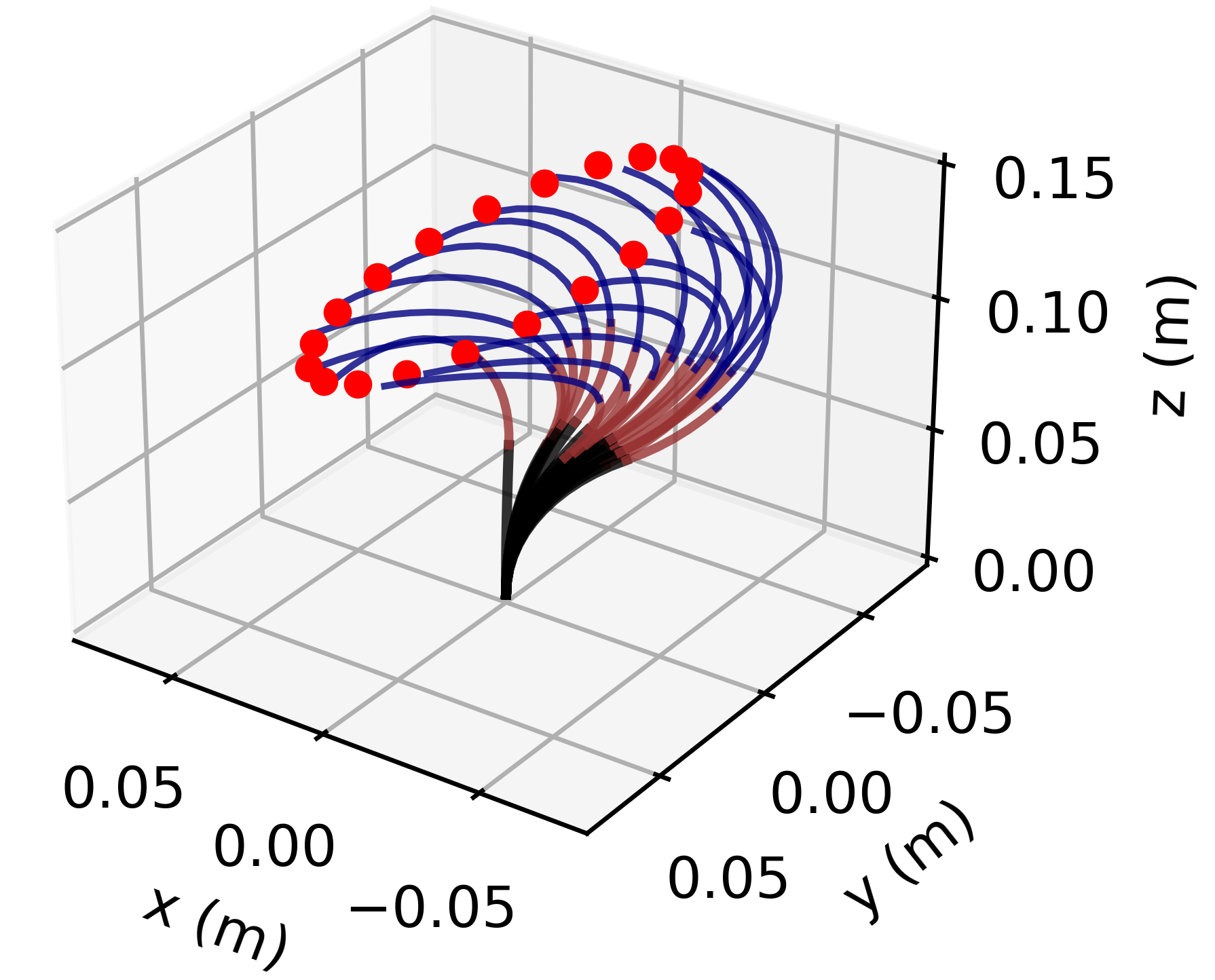}
    }
    \vspace{-2mm}
    \caption{Representative results for the trajectory tracking scenario: red dots shows the reference trajectories and colored solid lines indicate the three-segment robot: (a)~circle, (b)~square, (c)~S-shape, and (d)~elipse trajectories.}
    \vspace{-5mm}
    \label{fig:traj}    
\end{figure*}

\begin{figure*}[!t]
    \centering
    \subfigure[]{
        \includegraphics[width=0.23\textwidth]{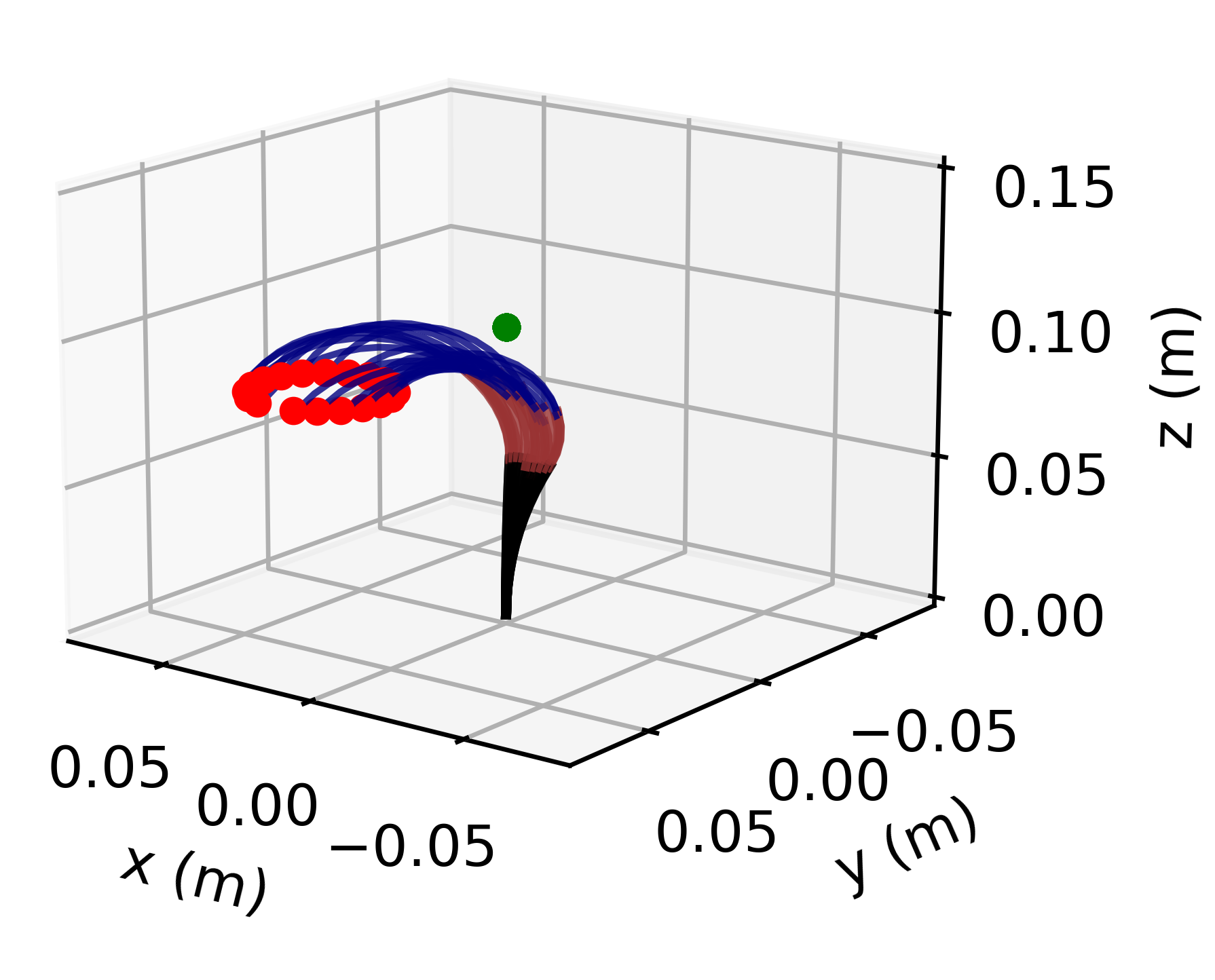}
    }
    \subfigure[]{
        \includegraphics[width=0.23\textwidth]{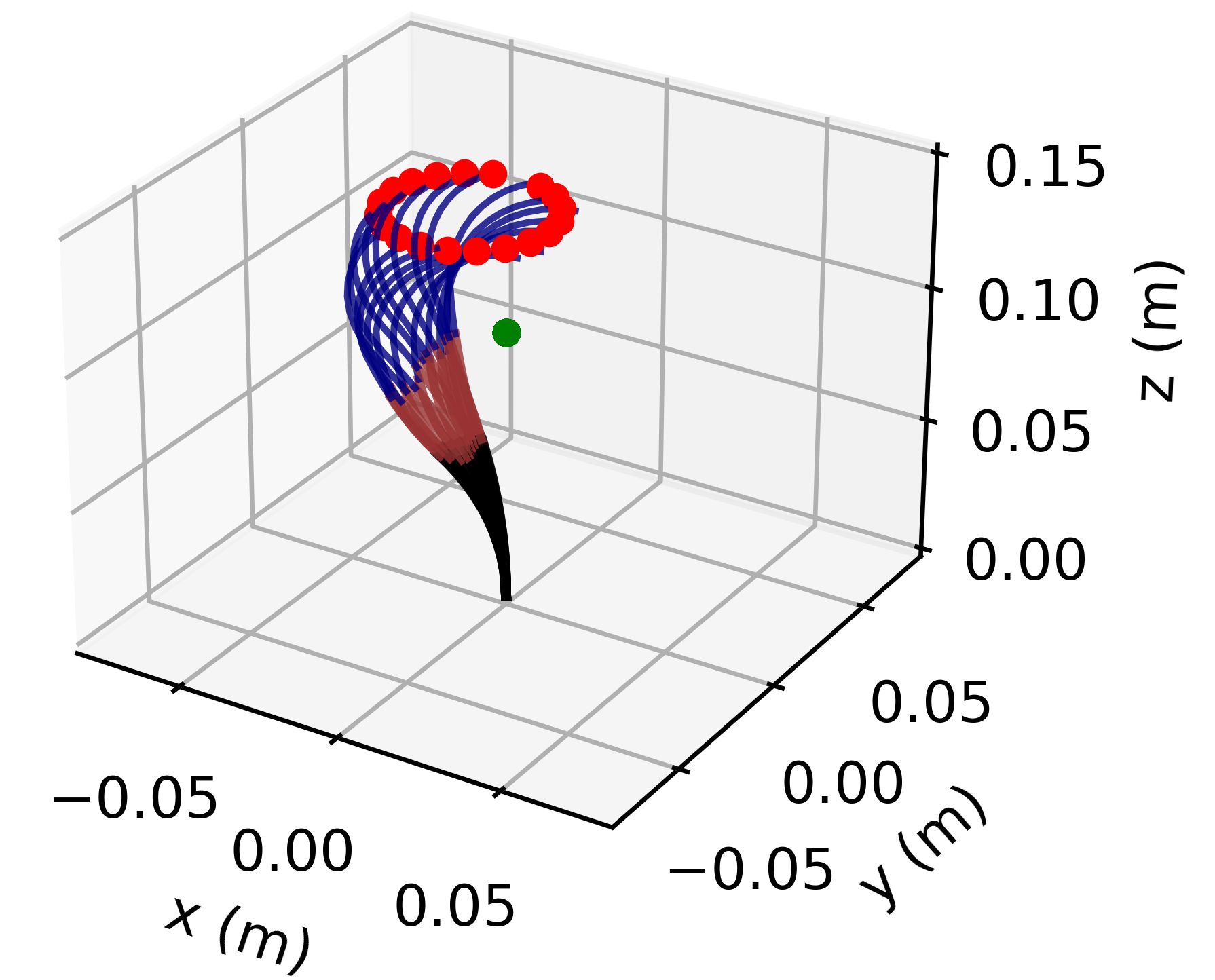}
    }
    \subfigure[]{
        \includegraphics[width=0.23\textwidth]{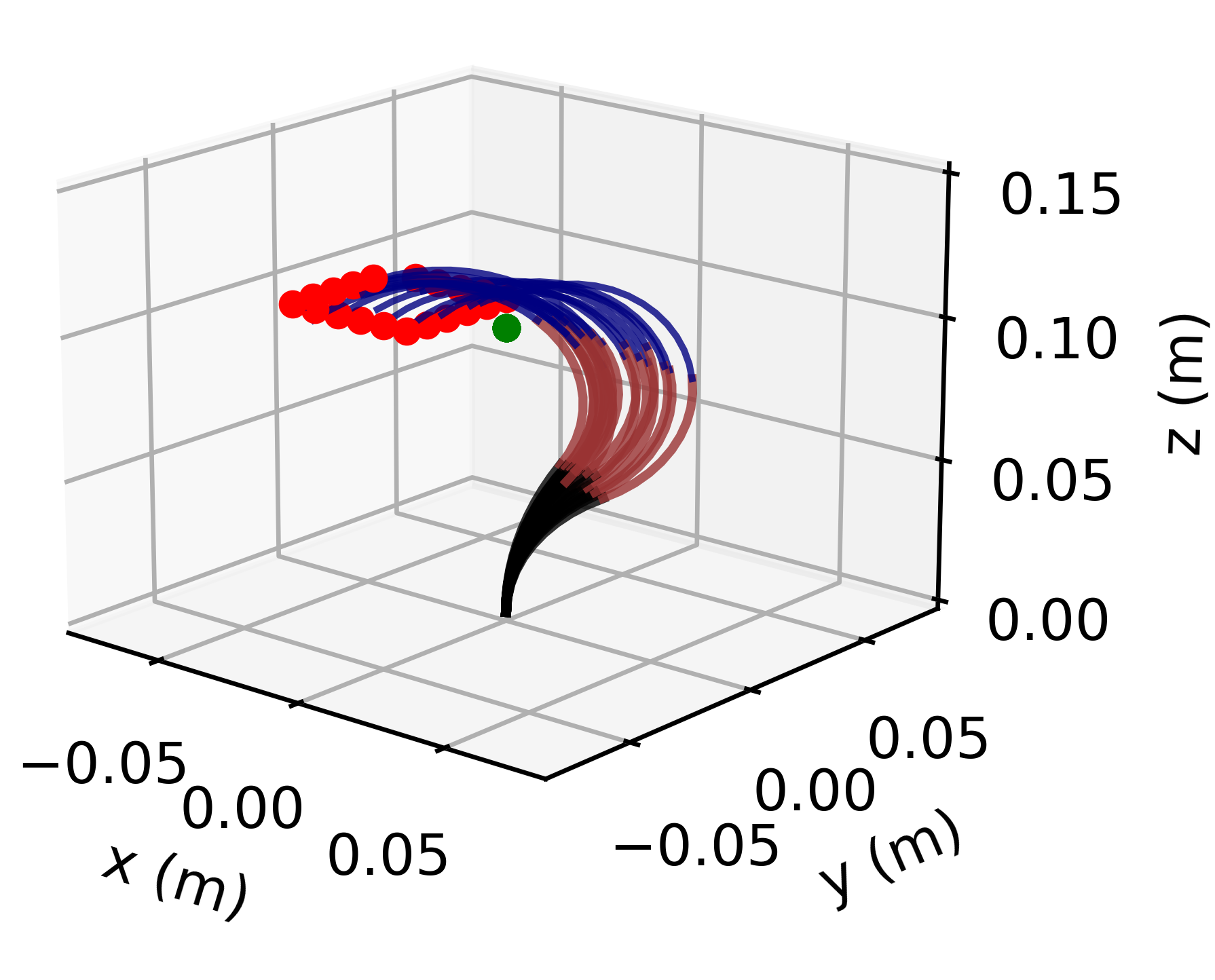}
    }
    \subfigure[]{
        \includegraphics[width=0.23\textwidth]{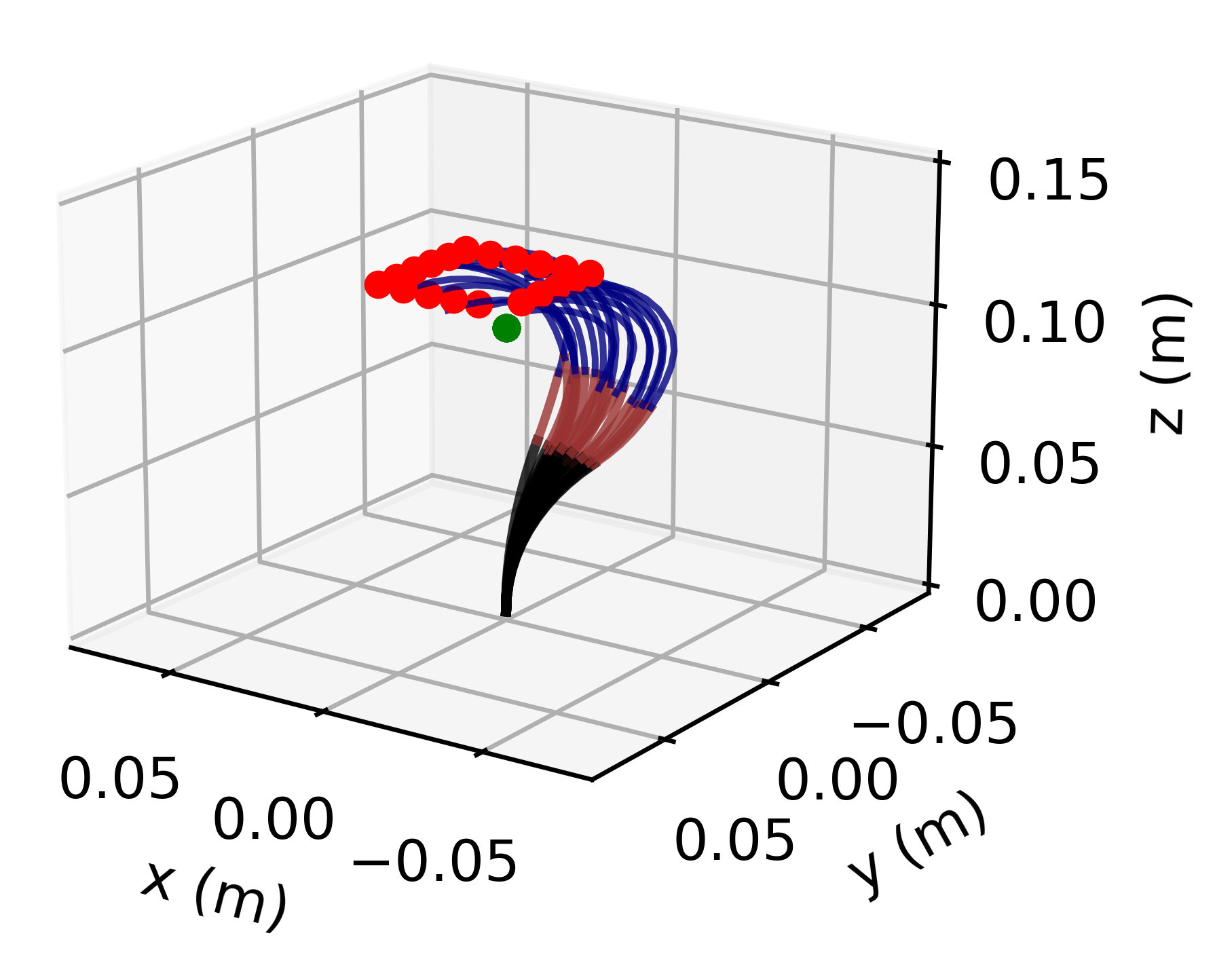}
    }
    \vspace{-2mm}
    \caption{Representative results for the obstacle avoidance scenario: red dots represent the reference trajectories, green dots indicate the position of the obstacle, and colored solid lines depict the path of the three-segment robot. The scenarios include: (a) a circular trajectory below the obstacle, (b) a circular trajectory above the obstacle, (c) a square trajectory near the obstacle, and (d) a square trajectory above the obstacle.}
    \label{fig:obs}
     \vspace{-5mm}

\end{figure*}

\subsection{Results and Discussions}

In the Shape Estimation scenario, 50 simulations were conducted for each robot, with representative results shown in Figure~\ref{fig:shape}. The performance of the shape prediction was evaluated by calculating the root mean squared error~(RMSE) and the standard deviation~(STD) of the error across the 50 trials, providing a quantitative measure of accuracy and variability. The results are summarized in Table~\ref{tab:shape_prediction}, demonstrate the effectiveness of the \textit{Shape-NODE} framework across different robot configurations. 
For the one- and two-segment robots, the RMSE remains relatively low across all axes, with values below 0.6 mm, indicating high prediction accuracy. As the number of segments increases, the RMSE and STD values show a noticeable rise, particularly for the three- and four-segment robots, with RMSE reaching as high as 2.161 mm for the y-axis in the four-segment case. This increase in error and variability suggests a growing challenge in accurately predicting more complex shapes with multiple segments, likely due to the increased degrees of freedom and non-linearities in the robot's shape. However, the overall performance remains robust, with relatively low STD values indicating consistent performance across the trials.

\begin{table}[!htb]
		\vspace{-3mm}
\caption{Shape Prediction Results}
		\begin{center}
\vspace{-3mm}
			\begin{tabular}{ |c | c | c | c | c | c | c |}
				\cline{2-7}
				\nocell{1} &  \multicolumn{3}{|c|}{RMSE} & \multicolumn{3}{|c|}{STD}  \\ \cline{2-7}
				\nocell{1} & \multicolumn{1}{|c|}{$\tilde{x}$} & $\tilde{y}$  & $\tilde{z}$ & \multicolumn{1}{|c|}{$\tilde{x}$} & $\tilde{y}$  & $\tilde{z}$\\    
				
				\nocell{1}& \multicolumn{1}{|c|}{(\SI{}{m\meter})} & (\SI{}{m\meter}) & (\SI{}{m\meter}) & \multicolumn{1}{|c|}{(\SI{}{m\meter})} & (\SI{}{m\meter}) & (\SI{}{m\meter})\\    \hline
				One-Segment   & \multicolumn{1}{|c|}{0.539} &  0.545  &  0.290 & \multicolumn{1}{|c|}{0.538} & 0.547  & 0.290\\    \hline 
				Two-Segments   & \multicolumn{1}{|c|}{0.314} & 0.349  & 0.357 & \multicolumn{1}{|c|}{0.314} & 0.347  & 0.357\\    \hline
			    Three-Segments   & \multicolumn{1}{|c|}{1.101} & 1.169  & 0.981 & \multicolumn{1}{|c|}{1.093} & 1.168  & 0.972\\    \hline
                    Four-Segments   & \multicolumn{1}{|c|}{1.917} & 2.161  & 2.094 & \multicolumn{1}{|c|}{1.913} & 2.160  & 2.094\\ \hline
                   
			\end{tabular}
		\end{center}
		\label{tab:shape_prediction}
		\vspace{-3mm}
\end{table}






The results of the Trajectory Tracking task, summarized in Table~\ref{tab:trajectory_tracking_table}, indicate the performance of the \textit{Control-NODE} across various trajectories. The RMSE values show that the S-shape trajectory had the lowest tracking error, particularly in the x-axis with an RMSE of 1.092 mm, demonstrating the highest accuracy in this case. In contrast, the square trajectory exhibited the highest RMSE in the x-axis at $6.059$~mm, indicating greater difficulty in accurately following sharp corners. The circular and elliptical trajectories presented moderate errors, with the circle showing higher RMSE in the x-axis~($4.432$~mm), while the ellipse had a more balanced error across all axes. The standard deviation~(STD) results reveal relatively low variability in all cases, indicating consistent tracking performance across the 5 tests for each trajectory. Overall, these results highlight that the \mbox{\textit{Control-NODE}} performs best with smooth trajectories like the S-shape, while it encounters more challenges with angular or sharp-cornered paths like the square trajectory. A set of representative results are depicted in Figure~\ref{fig:traj}.

\begin{table}[!t]
		\vspace{-3mm}
\caption{Trajectory Tracking Results}
		\begin{center}
\vspace{-3mm}
			\begin{tabular}{ |c | c | c | c | c | c | c |}
				\cline{2-7}
				\nocell{1} &  \multicolumn{3}{|c|}{RMSE} & \multicolumn{3}{|c|}{STD}  \\ \cline{2-7}
				\nocell{1} & \multicolumn{1}{|c|}{$\tilde{x}$} & $\tilde{y}$  & $\tilde{z}$ & \multicolumn{1}{|c|}{$\tilde{x}$} & $\tilde{y}$  & $\tilde{z}$\\    
				
				\nocell{1}& \multicolumn{1}{|c|}{(\SI{}{m\meter})} & (\SI{}{m\meter}) & (\SI{}{m\meter}) & \multicolumn{1}{|c|}{(\SI{}{m\meter})} & (\SI{}{m\meter}) & (\SI{}{m\meter})\\    \hline
  				Circle   & \multicolumn{1}{|c|}{4.432} & 2.621  & 1.77 & \multicolumn{1}{|c|}{2.259} & 1.840  & 1.051\\    \hline
      			Square   & \multicolumn{1}{|c|}{6.059} & 2.332  & 1.578 & \multicolumn{1}{|c|}{3.577} & 1.412  & 0.963\\    \hline
    
			    S-Shape   & \multicolumn{1}{|c|}{1.092} & 2.854  & 1.400 & \multicolumn{1}{|c|}{1.150} & 1.684  & 0.996\\    \hline
    
                    Elipse   & \multicolumn{1}{|c|}{5.176} & 2.822  & 0.998 & \multicolumn{1}{|c|}{2.584} & 1.779  & 0.724\\ \hline
                    
			\end{tabular}
		\end{center}
		\label{tab:trajectory_tracking_table}
		\vspace{-0mm}
\end{table}

To perform the Obstacle Avoidance scenario, a binary term was introduced into the loss function~(\ref{eq:loss}), \mbox{($100$~\texttt{torch.lt}$(\|{\mathbf{o}}(t_k) - \hat{\vect{p}}(S_{k-1})\|^2)$,~$0.01$)}, which penalizes the robot for approaching too close~(1cm) to the obstacle. A new policy was trained using this modified loss function. The trained policy allows a three-segment continuum robot to effectively follow both circular and square trajectories while avoiding a static obstacle. The representative results, depicted in Fig.~\ref{fig:obs}, demonstrate the effectiveness of the proposed framework.
The robot successfully navigates around the obstacle (green dot) without collision, maintaining smooth transitions in its shape. This is particularly evident in the circular trajectories~(a, b), where the robot curves gracefully around the obstacle, and in the square trajectories~(c, d), where the robot efficiently avoids the obstacle while still tracking the sharp turns of the square path. The framework proves capable of maintaining consistent robot shape and avoiding abrupt changes, even in the presence of an obstacle, showcasing its robustness in more challenging environments.

\section{Experiments}
\label{sec:exp}
Here, we conducted a series of experiments to evaluate the performance of the proposed method in different scenarios.

\begin{figure}[!t]
    \centering

    \subfigure[]{
        \includegraphics[width=0.45\linewidth]{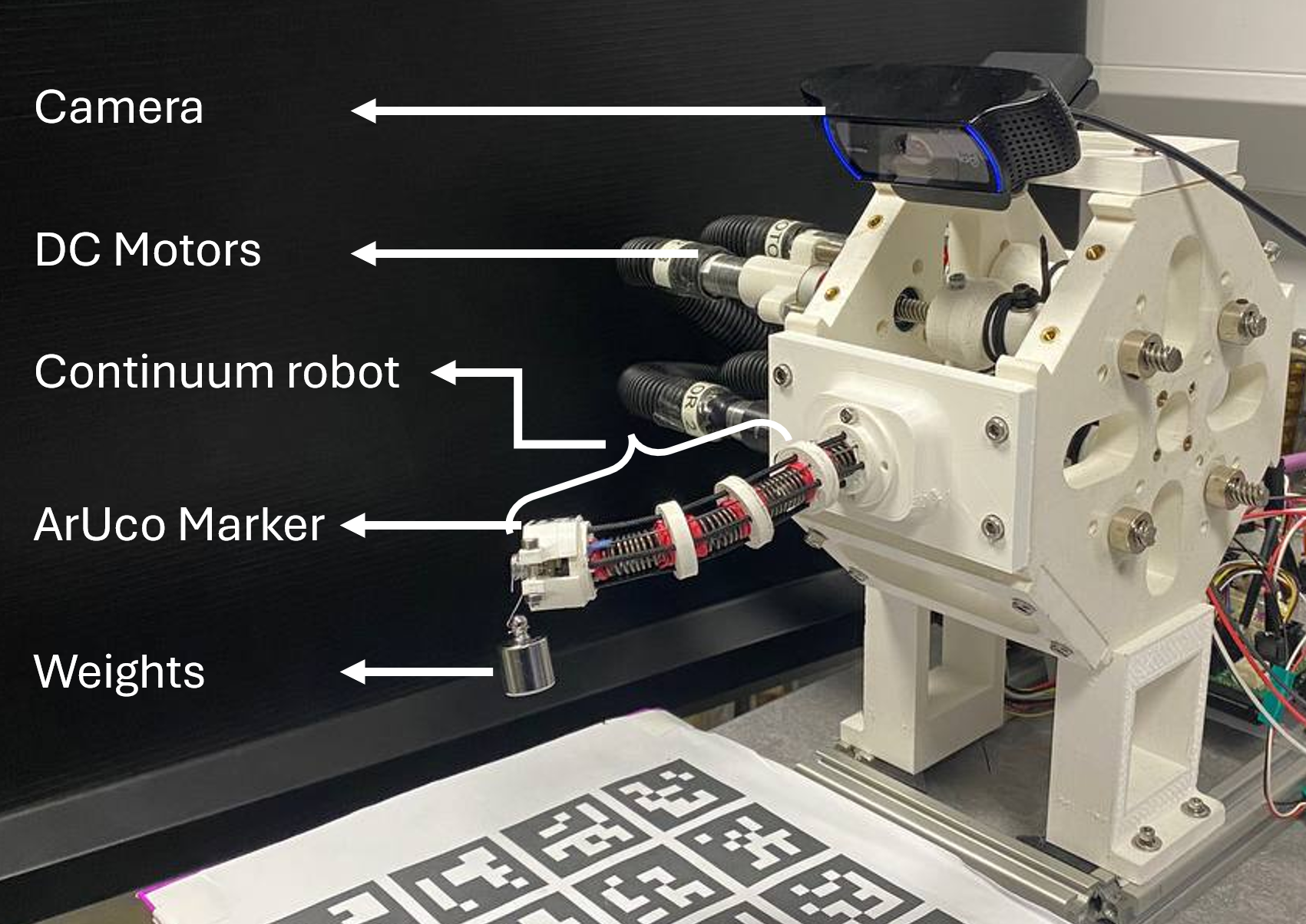}
        \label{fig:setup}
    }
    \subfigure[]{
        \includegraphics[trim={37.6cm 1cm 0cm 0},clip, width=0.404\linewidth]{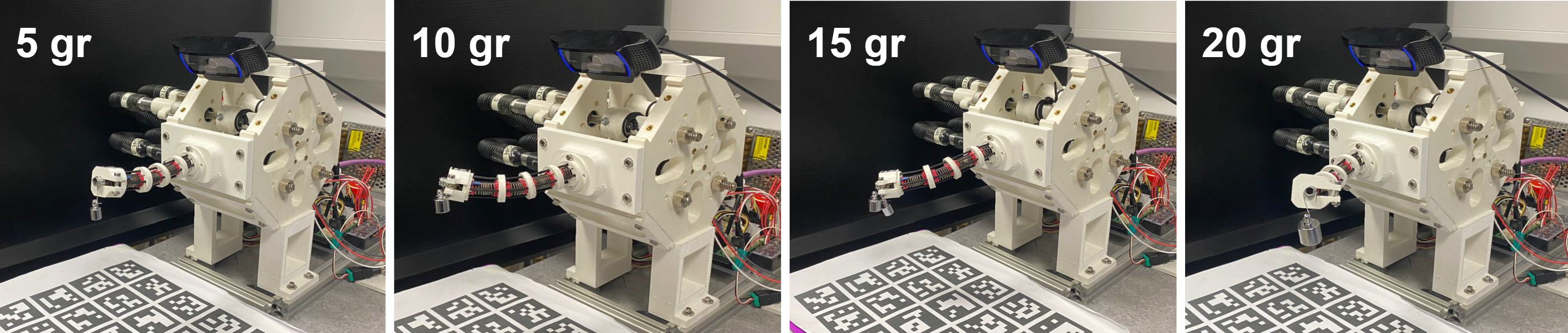}
        \label{fig:weight_real}
    }
    \vspace{-3mm}
    \caption{ (a)~Experiment setup; (b)~a snapshot showing the robot tracking the trajectory while carrying 20 grams weight.}
    \label{fig:weights}
    \vspace{-3mm}
\end{figure}

\subsection{Experiment Setup}
Figure~\ref{fig:setup} shows the robot used in our experiments, which consists of a flexible backbone supported by spacers.
A Logitech RGB camera is mounted at the robot's base, and an ArUco marker~\cite{Aruco2014, GarridoJurado2015} is attached to the tip of the robot for precise position tracking, forming a crucial part of the feedback loop.

\subsection{Experiments}
Two experimental scenarios were designed to evaluate the performance of the proposed framework on the real robot. The first scenario focuses on trajectory tracking, where the robot is tasked with following various trajectories, consistent with those used in the simulations. The second scenario is designed to assess the robustness of the method by requiring the robot to follow a helical trajectory while carrying additional payloads, with weights ranging from 5 to 20 grams (see Fig.~\ref{fig:weight_real}). A video demonstrating the experiments and results is available online.\footnote{\url{https://youtu.be/aZWfDQK9hpI?si=52BSPRC4p7HSqjsl}}

\begin{figure}[!t]
    \centering
    \subfigure[]{
        \includegraphics[width=0.44\linewidth]{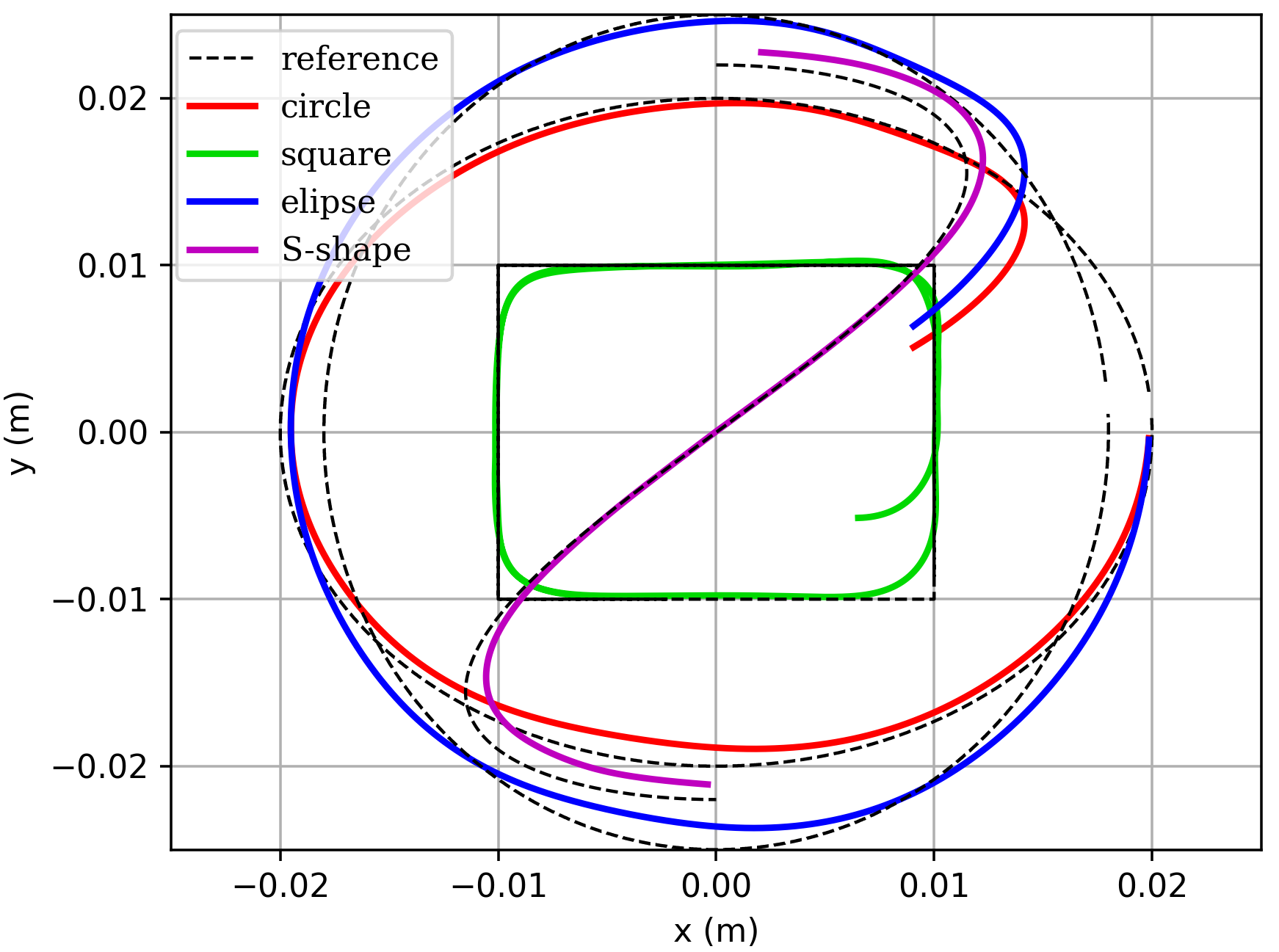}
        \label{fig:traj_real}
    }
    \subfigure[]{
        \includegraphics[width=0.46\linewidth]{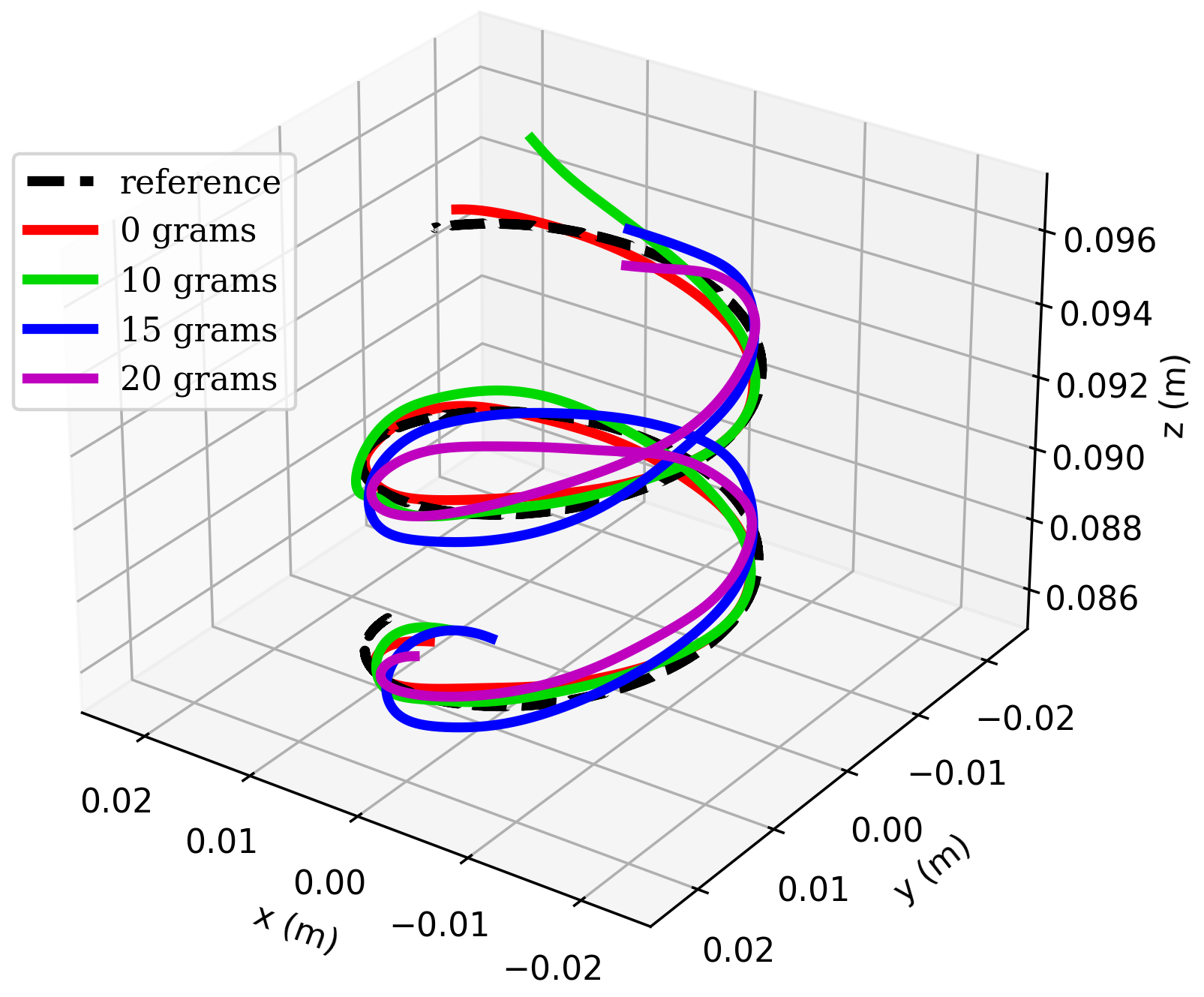}
        \label{fig:weight_real}
    }
    \caption{Real robot experiments results; (a)~representative results of trajectory tracking; (b)~the robot is set to follow a helical trajectory while carrying additional payloads, with weights ranging from 5 to 20 grams. }
    \vspace{-3mm}
    \label{fig:real_robot_res}    
\end{figure}

\subsection{Results and Discussions}
The results of the trajectory tracking experiments are summarized in Table~\ref{tab:trajectory_tracking_table_real_robot} and representative results are depicted in Fig.~\ref{fig:traj_real}. These results indicate that the robot's performance varies significantly depending on the trajectory shape. The smooth S-shape trajectory yields the lowest RMSE and STD values, suggesting that the robot can follow continuous curves with high accuracy and consistency. In contrast, the square trajectory, with its sharp turns, leads to the highest RMSE, particularly in the x and z directions, indicating that abrupt directional changes are more challenging for the control system. Circular and elliptical trajectories exhibit moderate performance, with slightly higher variability in the y-axis for the ellipse. Overall, the results demonstrate that the proposed framework performs best with smoother trajectories, while more complex shapes lead to increased error and variability.
\begin{table}[!t]
		\vspace{-3mm}
\caption{Real Robot Trajectory Tracking Results}
		\begin{center}
\vspace{-3mm}
			\begin{tabular}{ |c | c | c | c | c | c | c |}
				\cline{2-7}
				\nocell{1} &  \multicolumn{3}{|c|}{RMSE} & \multicolumn{3}{|c|}{STD}  \\ \cline{2-7}
				\nocell{1} & \multicolumn{1}{|c|}{$\tilde{x}$} & $\tilde{y}$  & $\tilde{z}$ & \multicolumn{1}{|c|}{$\tilde{x}$} & $\tilde{y}$  & $\tilde{z}$\\    
				
				\nocell{1}& \multicolumn{1}{|c|}{(\SI{}{m\meter})} & (\SI{}{m\meter}) & (\SI{}{m\meter}) & \multicolumn{1}{|c|}{(\SI{}{m\meter})} & (\SI{}{m\meter}) & (\SI{}{m\meter})\\    \hline
  				Circle   & \multicolumn{1}{|c|}{3.11} & 3.41  & 3.23 & \multicolumn{1}{|c|}{3.81} & 3.01  & 3.021\\    \hline
      			Square   & \multicolumn{1}{|c|}{5.04} & 4.31  & 4.81 & \multicolumn{1}{|c|}{2.61} & 3.72  & 3.01\\    \hline
    
			    S-Shape   & \multicolumn{1}{|c|}{2.12} & 2.22  & 1.83 & \multicolumn{1}{|c|}{2.01} & 2.04  & 1.86\\    \hline
    
                    Elipse   & \multicolumn{1}{|c|}{3.16} & 3.82  & 2.18 & \multicolumn{1}{|c|}{2.41} & 2.79  & 2.43\\ \hline
                    
			\end{tabular}
		\end{center}
		\label{tab:trajectory_tracking_table_real_robot}
		\vspace{-3mm}
\end{table}
The results depicted in Fig.~\ref{fig:weight_real}, showing the helical trajectory tracking under different payload conditions, demonstrate the robustness of the proposed framework with regarding to external loads. Despite the addition of varying payloads ranging from 0 to 20~g, the robot is able to maintain trajectory accuracy with minimal deviation from the reference. The results shows that the robot's path remains closely aligned with the reference trajectory across all payload conditions, with only slight deviations as the weight increases. This indicates that the control framework effectively compensates for the added load, maintaining stable performance and accurate tracking even when faced with unknown external disturbances, like varying payloads. Hence, the framework's adaptability to these changes showcases its robustness and reliability in real-world scenarios.

\section{Comparison Study}
\label{sec:comp}
In this section, we conducted a series of simulations to compare the performance of the framework against end-to-end models, Neural-ODE-based approaches, and Recurrent Neural Networks~(RNN) across various scenarios. Our comparison primarily focuses on two key aspects: trajectory tracking accuracy (open-loop and closed-loop) and generalization beyond the training data. 

In the open-loop scenario, we simplified the control strategy by excluding the \textit{Control-NODE} and instead used a basic feedforward approach. Specifically, we employed the equation $\dot{\vect{q}} = \mathbf{J}^{+}\dot{\vect{x}}$, where $\mathbf{J}^{+}$ is the pseudo-inverse of the Jacobian that can be calculated using \textit{Shape-Node} and $\dot{\vect{x}}$ represents the desired reference trajectory. While this approach achieved reasonable accuracy in the straighter sections, the absence of feedback led to increasing errors at the sharp turns. In contrast, in the closed-loop control scenario, the real-time feedback mechanism provided corrective inputs, allowing the \textit{Control-NODE} to refine its actions and handle the sharp corners, significantly reducing tracking errors ($2.653\pm1.737$ mm) compared to the open-loop setup~($9.265\pm1.063$ mm). When compared to RNN~($4.653\pm1.836$ mm), end-to-end~($6.595\pm2.058$ mm), and Neural ODE models~($3.052\pm1.078$ mm), our framework consistently outperformed them in both open-loop and closed-loop scenarios. The end-to-end and RNN models struggled particularly with the sharp turns of the square, resulting in larger tracking errors. Neural ODE models performed better than end-to-end approaches, but still exhibited less accuracy than the proposed framework.

Next, we evaluated the generalization capability of each model by testing them on unseen data (not included in the training set). The proposed framework, leveraging its ANODE-based structure, demonstrated strong generalization performance. The \textit{Shape-NODE}, which integrates prior knowledge from Cosserat rod theory, enabled the framework to  estimate the robot’s shape even in scenarios beyond the training distribution. Results showed that the framework maintained stable performance with only a slight increase in RMSE. In contrast, end-to-end models and RNNs experienced significant declines in accuracy when tested on new data, underscoring their limited generalization ability. Neural ODE models generalized better than the end-to-end models but still did not perform as well as the proposed framework.

\section{Conclusion}
\label{sec:conclusion}
This paper introduced a synergistic framework that combines shape estimation and shape-aware control for continuum robots using two Augmented Neural Ordinary Differential Equations~(ANODEs). The integration of the \textit{Shape-NODE}, with its built-in model of Cosserat rod theory, allowed for accurate shape prediction and adaptation to model mismatches, while the \textit{Control-NODE} enabled the development of a shape-aware whole-body control policy optimized in an MPC fashion. Our  simulations and experiments validate the effectiveness and robustness of the framework, especially in complex scenarios involving trajectory tracking and obstacle avoidance. The proposed framework consistently outperformed alternative approaches, like end-to-end models, Neural ODEs, and RNNs, both in terms of tracking accuracy and generalization to unseen data. 



\bibliographystyle{IEEEtran}
\bibliography{Main}  

\end{document}